\documentclass{article}
\usepackage{ijcai19}

\usepackage{times}
\usepackage{soul}
\usepackage{url}
\usepackage[hidelinks]{hyperref}
\usepackage[utf8]{inputenc}
\usepackage[small]{caption}
\usepackage{graphicx}
\usepackage{amsmath}
\usepackage{booktabs}

\usepackage{amssymb}
\usepackage{amsthm}
\usepackage{bm}
\usepackage{adjustbox}
\usepackage{multirow}
\usepackage{enumerate}

\newcommand{\bs}{\boldsymbol}

\makeatletter
\def\th@definition{
  \thm@notefont{}
  \normalfont
}
\makeatother
\theoremstyle{definition}
\newtheorem{definition}{Definition} 
\newcommand{\sigb}[1]{\textbf{#1}}
\newcommand{\sigu}[1]{\underline{#1}}

\title{Representation Learning with Weighted Inner Product\\ for Universal Approximation of General Similarities}

\author{
Geewook Kim\and
Akifumi Okuno\and
Kazuki Fukui\And
Hidetoshi Shimodaira\\
\affiliations
Department of Systems Science, Graduate School of Informatics, Kyoto University\\
Mathematical Statistics Team, RIKEN Center for Advanced Intelligence Project\\
\emails
\{geewook, okuno, k.fukui\}@sys.i.kyoto-u.ac.jp,
shimo@i.kyoto-u.ac.jp
}

\begin{document}

\maketitle

\begin{abstract}
We propose \emph{weighted inner product similarity} (WIPS) for neural network-based graph embedding. In addition to the parameters of neural networks, we optimize the weights of the inner product by allowing positive and negative values.
Despite its simplicity, WIPS can approximate arbitrary general similarities including positive definite, conditionally positive definite, and indefinite kernels.
WIPS is free from similarity model selection, since it can learn any similarity models such as cosine similarity, negative Poincar\'e  distance and negative Wasserstein distance.
Our experiments show that the proposed method can learn high-quality distributed representations of nodes from real datasets, leading to an accurate approximation of similarities as well as high performance in inductive tasks.
\end{abstract}

\section{Introduction}
Representation learning of graphs, also known as graph embedding, computes vector representations of nodes in graph-structured data.
The learned representations called \emph{feature vectors} are widely-used in a range of applications, e.g., community detection or link prediction on social networks \cite{perozzi2014deepwalk,Tang:2015:LLI:2736277.2741093,Hamilton2017InductiveRL}.
Words in a text corpus also constitute a co-occurrence graph, and the learned feature vectors of words are successfully applied in many natural language processing tasks \cite{Mikolov:2013:DRW:2999792.2999959,glove,Goldberg:2016:PNN:3176748.3176757}.

The feature vector is a model parameter or computed from the node's attributes called \emph{data vector}.
Many of state-of-the-art graph embedding methods train feature vectors in a low dimensional Euclidean space so that their inner product expresses the similarities representing the strengths of association among nodes.
We call this common framework as \textit{inner product similarity~(IPS)} in this paper.
IPS equipped with a sufficiently large neural network is highly expressive and it has been proved to approximate arbitrary Positive Definite (PD) similarities~\cite{DBLP:conf/icml/OkunoHS18}, e.g., cosine similarity.
However, IPS cannot express non-PD similarities.

To express some non-PD similarities, similarity functions based on specific kernels other than inner product have also been considered.
For instance, Nickel et al.~\shortcite{NIPS2017_7213poincare} applied Poincar\'e distance to learn a hierarchical structure of words in hyperbolic space.
Whereas practitioners are required to specify a latent similarity model for representation learning on a given data, finding the best model is not a simple problem, leading to a tedious trial and error process. 
To address the issue, we need a highly expressive similarity model.

Recently, a simple similarity model named \textit{shifted inner product similarity~(SIPS)} has been proposed by introducing bias terms in IPS model~\cite{DBLP:journals/corr/abs-1810-03463}.
SIPS can approximate arbitrary conditionally PD~(CPD) similarities, which includes any PD similarities and some interesting similarities, e.g., negative Poincar\'e distance~\cite{NIPS2017_7213poincare} and negative Wasserstein distance~\cite{7780937}.
However, the approximation capability of SIPS is still limited to CPD similarities.

To further improve the approximation capability, \textit{inner product difference similarity~(IPDS)} has also been proposed by taking the difference between two IPS models~\cite[See Supplement E]{DBLP:journals/corr/abs-1810-03463}.
Any general similarities including indefinite kernels can be expressed as the difference between two PD similarities~\cite{ong2004learning}, thus they can be approximated by IPDS arbitrarily well.
However, only the concept of IPDS has been shown in the literature~\cite{DBLP:journals/corr/abs-1810-03463} without any experiments.

\begin{figure}[htbp]
\centering
\includegraphics[width=\linewidth]{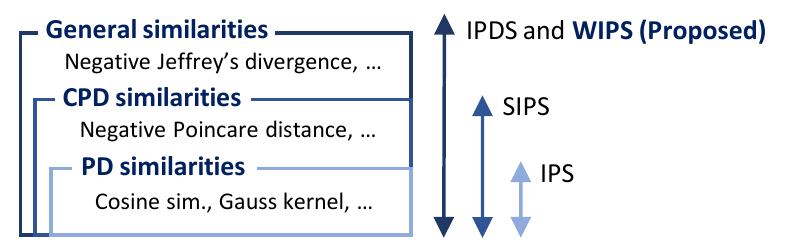}
\caption{Proposed WIPS is capable of approximating any general similarities, including PD, CPD and a variety of other similarities.}
\end{figure}

In this paper, we first examine IPDS on a range of applications to investigate its effectiveness and weakness.
There are, in fact, several practical concerns in IPDS such as a specification of dimensionalities of the two IPS models.
Then, in order to remedy the weakness of IPDS, we propose a new model named \textit{weighted inner product similarity (WIPS)}.
The core idea of WIPS is to incorporate element-wise weights to IPS so that it generalizes all the above-explained similarity models.
WIPS reduces to IPS when all the weights are $+1$, and it also reduces to IPDS when they are $+1$ or $-1$.
The weight values are optimized continuously by allowing positive and negative values, and thus WIPS is free from model selection, except for the choice of dimensionality.
Through our extensive experiments, we verify the efficacy of WIPS in real-world applications.

The main contributions of this work are: 
(i)~A new model WIPS is proposed for a universal approximation of general similarities without model selection.
(ii)~Experiments of WIPS, as well as IPDS, are conducted for showing their high approximation capabilities and effectiveness on a range of applications.
(iii)~Our code is publicly available at GitHub\footnote{\url{https://github.com/kdrl/WIPS}}.

\section{Related Works}
In this section, we overview some existing methods concerned with similarity learning.

\textbf{Metric learning} is a kind of similarity learning where similarity between two data vectors $\bs{x}_i, \bs{x}_j$ is captured by some metric function $d(\bs{x_i}, \bs{x_j})$~\cite{bellet2013survey,kulis2013metric}. A vast amount of previous works focused on linear metric learning, where Mahalanobis distance $d_{\bs{M}}(\bs{x_i}, \bs{x_j}) = \sqrt{(\bs x_i - \bs x_j)^\top \bs M (\bs x_i - \bs x_j)}$ is trained with respect to positive-semidefinite (PSD) matrix $\bs M$.
Finding such a PSD matrix $\bs M$ in this context is interpreted as finding a linear projection $\bs L$ that transforms each data vector by $\bs L^{\top} \bs x$~\cite{NIPS2004_2566}, since the PSD matrix $\bs M$ can be decomposed as $\bs{M} =\bs{L}\bs{L}^{\top}$. Therefore, it can also be seen as learning linear embedding so that similar feature vectors get closer.
It is extended to learning non-linear transformation by replacing $\bs L$ with neural networks, which has been known as Siamese architecture~\cite{bromley1994signature}.
When it comes to just capturing the similarities, especially with bilinear form of $\bs{x}^\top_i \bs{M} \bs{x}_j$, $\bs{M}$ no longer needs to be PSD~\cite{Chechik:2009:OAL:2984093.2984128}; these non-metric learning methods are closely related to our non-PD similarities, but they do not learn embedding representations of data vectors.

\textbf{Kernel methods} represent similarity between two data vectors by a positive-definite (PD) kernel.
Although the similarity can be shown as the inner product associated with Reproducing Kernel Hilbert Space~(RKHS), explicit embedding representations of data vectors are avoided via the kernel trick.
Kernel methods have been extended to conditionally PD (CPD) kernels~\cite{scholkopf2001kernel} for representing negative distances, and also to general kernels including indefinite kernels~\cite{ong2004learning,pmlr-v80-oglic18a} for representing a variety of non-PD kernels in real-world situations~\cite{Schleif:2015:IPL:2845878.2845880}.
The general kernels are theoretically expressed as inner products associated with Reproducing Kernel Kr\u{e}in Space~(RKKS), but such representations in infinite dimensions are avoided in practice.

\textbf{Graph embedding} can be interpreted as learning representations of nodes that preserve similarities between nodes defined from their links or neighbourhood structure.
The feature vector of a node is computed from its data vector using linear transformation~\cite{Yan:2007:GEE:1191551.1191800} or non-linear transformation~\cite{Wang:2016:SDN:2939672.2939753}, whereas feature vectors are treated as model parameters when data vectors are not observed~\cite{Tang:2015:LLI:2736277.2741093}.
Word embedding with Skip-gram model~\cite{Mikolov:2013:DRW:2999792.2999959} in natural language processing can also be interpreted as graph embedding without data vectors. They are often 2-view graphs, where vector representations of words and those of contexts are parameterized separately. Some graph embedding methods implement such 2-view settings~\cite{Tang:2015:LLI:2736277.2741093}, and its generalization to a multi-view setting is straightforward~\cite{DBLP:conf/icml/OkunoHS18}. Interestingly, most of these graph embedding models employ the inner product for computing similarity, thus they can be covered by our argument, although we work only on a 1-view case for simplicity.
Note that graph convolution networks~\cite{Hamilton2017InductiveRL} compute the feature vector of a node from the data vectors of its neighbour nodes, so they cannot be covered by our simple setting. However, these methods also use the inner product for computing similarity, hence, they may be improved by using our similarity models.

\section{Representation Learning for Graph Embedding}

Let us consider an undirected graph consisting of $n (\in \mathbb{N})$ nodes $\{v_i\}_{i=1}^{n}$ with a weighted adjacency matrix $(w_{ij}) \in \mathbb{R}_{\geq 0}^{n \times n}$. 
The symmetric weight $w_{ij}=w_{ji} \geq 0$ is called \emph{link weight}, that represents the observed strength of association between a node pair $(v_i,v_j)$ for all $1 \leq i < j \leq n$.
We may also consider a \emph{data vector} $\bs x_i \in \mathcal{X}$ at $v_i$, that takes a value in a set $\mathcal{X} \subset \mathbb{R}^p \: (p \in \mathbb{N})$.
The data vector represents attributes or side-information of the node.
If we do not have any such attributes, we use \emph{1-hot vector} instead, i.e., $\bs x_i$ is $n$-dimensional vector whose $i$-th entry is $1$ and $0$ otherwise, for $i=1,\ldots,n$. 
Our data consists of $\{w_{ij}\}_{i,j=1}^n$ and $\{\bs x_i\}_{i=1}^n$.

Following the setting of graph embedding described in \cite{DBLP:conf/icml/OkunoHS18,DBLP:journals/corr/abs-1810-03463},
we consider a \emph{feature vector} $\bs y_i\in\mathcal{Y}$ at $v_i$, that takes a value in a set $\mathcal{Y} \subset \mathbb{R}^K$ for some dimensionality $K\in\mathbb{N}$. Feature vectors are computed by a continuous transformation $\bs f: \mathcal{X}\to\mathcal{Y}$ as
\[
    \bs y_i:=\bs f(\bs x_i) \quad (i=1,2,\ldots,n).
\]
For two data vectors $\bs x, \bs x' \in \mathcal{X}$, the strength of association between $\bs x$ and $\bs x'$ is modeled by a \emph{similarity} function
\[
h(\bs x, \bs x') := g(\bs f(\bs x), \bs f(\bs x')),
\]
where $g: \mathcal{Y}^2 \to \mathbb{R}$ is a symmetric continuous function, e.g., the inner product $g(\bs y, \bs y') = \langle \bs y, \bs y'\rangle = \sum_{k=1}^K y_k y_k'$.
We employ a simple random graph model by specifying the conditional expectation of $w_{ij}$ given $\bs x_i$ and $\bs x_j$ as
\[
 E(w_{ij} | \bs x_i, \bs x_j) = \nu( h(\bs x, \bs x')),
\]
where $\nu: \mathbb{R}\to \mathbb{R}$ is a non-linear function;
e.g., $\nu(x) = \sigma(x) := (1+\exp(-x))^{-1}$ is a sigmoid function when $w_{ij}\in\{0,1\}$ follows Bernoulli distribution, or $\nu(x)=\exp(x)$ when $w_{ij}\in\{0,1,\ldots\}$ follows Poisson distribution.

For learning distributed representations of nodes from the observed data, we consider a parametric model $\bs f_{\bs \theta}:\mathcal{X} \to \mathcal{Y}$ with unknown parameter vector $\bs \theta \in \bs \Theta$.
Typically, $\bs f_{\bs \theta}$ is a non-linear transformation implemented as a vector valued neural network (NN), or a linear transformation. The similarity function $h$ is now modelled as
\begin{align}
h(\bs x,\bs x';\bs \theta, \bs \lambda):=g_{\bs \lambda}(\bs f_{\bs \theta}(\bs x),\bs f_{\bs \theta}(\bs x') ),
\label{eq:parametrized_similarity}
\end{align}
where $g_{\bs \lambda}(\bs y, \bs y')$ has also a parameter vector $\bs \lambda$ in general.
This architecture for similarity learning is called Siamese network~\cite{bromley1994signature}.
The parameter  $\bs \theta \in \bs \Theta$ as well as $\bs \lambda$ may be estimated by maximizing the log-likelihood function.
For the case of Bernoulli distribution (i.e., the binary $w_{ij}$), the objective function is
\begin{align}
    &\sum_{1 \le i<j \le n}
    w_{ij} \log \nu(h(\bs x_i,\bs x_j;\bs \theta, \bs \lambda)) \nonumber \\
    & \quad +
    \sum_{1 \le i<j \le n}
    (1-w_{ij}) \log (1-\nu(h(\bs x_i,\bs x_j;\bs \theta, \bs \lambda))),
    \label{eq:logistic_loss}
\end{align}
which is optimized efficiently by mini-batch SGD with negative sampling~\cite{DBLP:conf/icml/OkunoHS18}.
Once the optimal parameters $\hat{\bs \theta}$ and $\hat{\bs\lambda}$ are obtained, we compute feature vectors as
$ \bs y_i =\bs f_{\hat{\bs \theta}}(\bs x_i) $, $i=1,\ldots,n$.
For inductive tasks with newly observed data vectors $\bs x_i$ ($i=n+1,n+2,\ldots$), we also compute feature vectors by $\bs y_i =\bs f_{\hat{\bs \theta}}(\bs x_i)$ unless $\bs x_i$ are 1-hot vectors, and then we can compute the similarity function $g_{\hat{\bs \lambda}}(\bs y_i,\bs y_j)$ for any pair of feature vectors.

\section{Existing Similarity Models and Their Approximation Capabilities}\label{sec:similarities}

The quality of representation learning relies on the model (\ref{eq:parametrized_similarity}) of similarity function.
Although arbitrary similarity functions can be expressed if $g_{\bs \lambda}$ is a large neural network with many units,
our $g_{\bs \lambda}$ are confined to simple extensions of inner product in this paper.
All the models reviewed in this section are very simple and they do not have the parameter $\bs \lambda$.
You will see very subtle extensions to the ordinary inner product greatly enhance the approximation capability.

\subsection{Inner Product Similarity (IPS)}
\label{subsec:ips}
Many of conventional representation learning methods use IPS as its similarity model. IPS is defined by
\begin{align*}
h_{\text{IPS}}(\bs x, \bs x';\bs \theta) 
 &= \langle \bs f_{\bs \theta}(\bs x),\bs f_{\bs \theta}(\bs x') \rangle,    
\end{align*}
where $\langle \bs y,\bs y' \rangle=\sum_{k=1}^K y_k y_k'$ represents the inner product in Euclidean space $\mathbb{R}^K$.
Note that, $h_{\text{IPS}}(\bs x,\bs x';\bs \theta)$ reduces to Mahalanobis inner product $\bs x^{\top}\bs L_{\bs \theta}\bs L_{\bs \theta}^{\top}\bs x'$~\cite{kung2014kernel},
by specifying $\bs f_{\bs \theta}(\bs x)$ as a linear-transformation $\bs L_{\bs \theta}^{\top}\bs x$ with a matrix $\bs L_{\bs \theta}\in \mathbb{R}^{p \times K}$ parameterized by $\bs \theta \in \bs \Theta$. 

For describing the approximation capability of IPS, we need the following definition.
The similarity function $h$ is also said as a kernel function in our mathematical argument.

\begin{definition}[Positive definite kernel]
A symmetric function $h: \mathcal{X}^2\to\mathcal{R}$ is said to be \emph{positive-definite~(PD)} if $\sum_{i=1}^{n}\sum_{j=1}^{n}c_i c_j h(\bs x_i,\bs x_j) \ge 0$ for any $\{\bs x_i\}_{i=1}^{n} \subset \mathcal{X}$ and $\{c_i\}_{i=1}^{n} \subset \mathbb{R}$. 
This definition of PD includes positive semi-definite.
Note that $h$ is called negative definite when $-h$ is positive definite.
\end{definition}

For instance, $h(\bs x, \bs x') = g(\bs f(\bs x), \bs f(\bs x'))$ is PD when $g(\bs y, \bs y') = \langle \bs y, \bs y'\rangle$ (IPS), and
$g(\bs y, \bs y') = \langle \bs y, \bs y'\rangle / \|\bs y\|_2 \|\bs y'\|_2 $ (cosine similarity).

IPS using a sufficiently large NN has been proved to approximate \emph{arbitrary} PD similarities~\cite[Theorem~3.2]{DBLP:conf/icml/OkunoHS18}, in the sense that, 
for any PD-similarity $h_*$, 
$h_{\text{IPS}}(\bs x,\bs x';\hat{\bs \theta})$ equipped with a NN converges to $h_*(\bs x,\bs x')$ as the number of units in the NN and the dimension $K$ increase.
The theorem is proved by combining the universal approximation theorem of NN 
\cite{funahashi1989approximate,cybenko1989approximation,yarotsky2016error,pmlr-v70-telgarsky17a},
and the Mercer's theorem
\cite{minh2006mercer}.
Therefore, IPS, despite its simplicity, covers a very large class of similarities.
However, IPS cannot approximate non-PD similarities  \cite{DBLP:journals/corr/abs-1810-03463}, such as negative Poincar\'e distance~\cite{NIPS2017_7213poincare} or
negative-squared distance $h(\bs x, \bs x') = -\|\bs f(\bs x)- \bs f(\bs x')\|_2^2$.

\subsection{Shifted Inner Product Similarity (SIPS)}
\label{subsec:sips}
For enhancing the IPS's approximation capability, Okuno et al.~\shortcite{DBLP:journals/corr/abs-1810-03463} has proposed \emph{shifted inner product similarity~(SIPS)} defined by
\begin{align*}
h_{\text{SIPS}}(\bs x, \bs x';\bs \theta) 
 &= \langle \tilde{\bs f}_{\bs \theta}(\bs x),\tilde{\bs f}_{\bs \theta}(\bs x') \rangle + 
 u_{\bs \theta}(\bs x) + u_{\bs \theta}(\bs x'),    
\end{align*}
where $\tilde{\bs f}_{\bs \theta}:\mathcal{X} \to \mathbb{R}^{K-1}$ and $u_{\bs \theta}:\mathcal{X} \to \mathbb{R}$
are continuous functions parameterized by $\bs \theta$. 
Thus $\bs f_{\bs \theta}(\bs x) = (\tilde{\bs y}, z) := (\tilde{\bs f}_{\bs \theta}(\bs x), u_{\bs \theta}(\bs x)) \in \mathbb{R}^K$ represents the feature vector for $\bs x$. 
By introducing the bias terms $ u_{\bs \theta}(\bs x) + u_{\bs \theta}(\bs x')$, SIPS extends the approximation capability of IPS.

For describing the approximation capability of SIPS, we need the following definition.

\begin{definition}[Conditionally positive definite kernel]
A symmetric function $h: \mathcal{X}^2\to\mathcal{R}$ is said to be \emph{conditionally PD~(CPD)} if $\sum_{i=1}^{n}\sum_{j=1}^{n}c_i c_j h(\bs x_i,\bs x_j) \ge 0$ for any $\{\bs x_i\}_{i=1}^{n} \subset \mathcal{X}$ and $\{c_i\}_{i=1}^{n} \subset \mathbb{R}$ satisfying $\sum_{i=1}^{n} c_i=0$. 
\end{definition}

CPD includes many examples, such as any of PD similarities, negative squared distance, negative Poincar\'e distance used in Poincar\'e embedding~\cite{NIPS2017_7213poincare} and some cases of negative Wasserstein distance~\cite{DBLP:conf/nips/XuWLC18}.

SIPS using a sufficiently large NN has been proved to approximate \emph{arbitrary} CPD similarities~\cite[Theorem~4.1]{DBLP:conf/icml/OkunoHS18}.
Therefore, SIPS covers even larger class of similarities than IPS does.
However, there still remain non-CPD similarities \cite[Supplement~E.3]{DBLP:conf/icml/OkunoHS18} such as negative Jeffrey's divergence and Epanechnikov kernel.

\subsection{Inner Product Difference Similarity (IPDS)}
\label{subsec:ipds}
For further enhancing the SIPS's approximation capability, Okuno et al.~\shortcite{DBLP:journals/corr/abs-1810-03463} has also proposed \textit{inner-product difference similarity~(IPDS)}.
By taking the difference between two IPSs, we define IPDS by
\begin{align*}
h_{\text{IPDS}}(\bs x, \bs x';\bs \theta)
&= \langle \bs f^{+}_{\bs \theta}(\bs x),\bs f^{+}_{\bs \theta}(\bs x') \rangle \\
&\hspace{4em} - \langle \bs f^{-}_{\bs \theta}(\bs x),\bs f^{-}_{\bs \theta}(\bs x') \rangle,    
\end{align*}
where $\bs f^{+}_{\bs \theta}:\mathcal{X} \to \mathbb{R}^{K-q}$ and $\bs f^{-}_{\bs \theta}:\mathcal{X} \to \mathbb{R}^{q}$ ($0\le q \le K$) are continuous functions parameterized by $\bs\theta\in\bs\Theta$. Thus $\bs f_{\bs \theta}(\bs x) = (\bs y^{+}, \bs y^{-}):=(\bs f^{+}_{\bs \theta}(\bs x),f^{-}_{\bs \theta}(\bs x)) \in \mathbb{R}^K$ represents the feature vector for $\bs x$. 
SIPS with $K$ dimensions is expressed as IPDS with $K+2$ dimensions by specifying $\bs f^+(\bs x)=(\tilde{\bs f}_{\bs \theta}(\bs x),u_{\bs \theta}(\bs x),1)\in\mathbb{R}^{K+1}$ and $\bs f^-(\bs x)=u_{\bs \theta}(\bs x)-1\in\mathbb{R}$. 
Therefore, the class of similarity functions of IPDS includes that of SIPS.
Also note that IPDS with $q=1$ and a constraint $h_{\text{IPDS}}(\bs x, \bs x;\bs \theta) = -1 $ gives the bilinear form in the hyperbolic space, which is used as another representation for Poincar\'e embedding \cite{pmlr-v80-nickel18alorentz,leimeister2018skipgram}.
For vectors $\bs y_i=(\bs y_i^+,\bs y_i^-)$, the bilinear-form $\langle \bs y_i^+,\bs y_j^+ \rangle - \langle \bs y_i^-,\bs y_j^- \rangle$ is known as an inner product associated with a pseudo-Euclidean space~\cite{greub1975linear}, which is also said to be an indefinite inner product space, and such inner products have been used in applications such as pattern recognition~\cite{goldfarb1984unified}.

For describing the approximation capability of IPDS, we need the following definition.

\begin{definition}[Indefinite kernel]
A symmetric function $h: \mathcal{X}^2\to\mathcal{R}$ is said to be \emph{indefinite} if neither of $h$ nor $-h$ is positive definite. We only consider $h$  which satisfies the condition
\begin{align*}
 h_1 = h_2 +h \text{ is PD for some PD kernel }h_2,
\end{align*}
so that $h$ can be decomposed as $h=h_1-h_2$ with two PD kernels $h_1$ and $h_2$ \cite[Proposition~7]{ong2004learning}.
\end{definition}

We call $h$ is \emph{general similarity} function when $h$ is positive definite, negative definite, or indefinite.

IPDS using a sufficiently large NN (with many units, large $K-q$ and $q$) has been proved to approximate \emph{arbitrary} general similarities~\cite[Theorem~E.1]{DBLP:conf/icml/OkunoHS18}, thus IPDS should cover an ultimately large class of similarities.

\subsection{Practical Issues of IPDS}

Although IPDS can universally approximate any general similarities with its high approximation capability, only the theory of IPDS without any experiments nor applications has been presented in previous research~\cite{DBLP:journals/corr/abs-1810-03463}.
There are several practical concerns when IPDS is implemented.
The optimization of $\bs f_{\bs \theta}$ for IPDS can be harder than IPS or SIPS without proper regularization, because the expression of IPDS is redundant; for arbitrary $\bs f_{\bs \theta}^0$, replacing $\bs f_{\bs \theta}^\pm(\bs x)$ with $(\bs f_{\bs \theta}^\pm(\bs x),\bs f_{\bs \theta}^0(\bs x))$, respectively, results in exactly the same similarity model but with redundant dimensions.
Choice of the two dimensionalities, i.e., $K-q$ and $q$, should be important.
This leads to our development of new similarity model presented in the next section.

\section{Proposed Similarity Model}

We have seen several similarity models which can be used as a model of (\ref{eq:parametrized_similarity}) for the representation learning in graph embedding. In particular, IPDS has been mathematically proved to approximate any general similarities, but the choice of the two dimensionalities can be a practical difficulty.
In order to improve the optimization of IPDS, we propose a new similarity model by modifying IPDS slightly.
Our new model has a tunable parameter vector $\bs \lambda$ in $g_{\bs \lambda}$, whereas all the models reviewed in the previous section do not have the parameter $\bs \lambda$.
Everything is simple and looks trivial, yet it works very well in practice.

\subsection{Weighted Inner Product Similarity~(WIPS)}

We first introduce the inner product with tunable weights.

\begin{definition}[Weighted inner product]
For two vectors $\bs y=(y_1,y_2,\ldots,y_K)$, $\bs y'=(y'_1,y'_2,\ldots,y'_K) \in \mathbb{R}^K$, \emph{weighted inner product} (WIP) equipped with the weight vector $\bs \lambda=(\lambda_1,\lambda_2,\ldots,\lambda_K) \in \mathbb{R}^K$ is defined as
\begin{align*}
\langle \bs y,\bs y' \rangle_{\bs \lambda}:=\sum_{k=1}^{K}\lambda_k y_k y_k'.
\end{align*}
The weights $\{\lambda_k\}_{k=1}^{K}$ may take both positive and negative values in our setting; thus, WIP is an indefinite inner product~\cite{bottcher1996lectures}. 
\end{definition}

WIP reduces to the ordinary inner product $\langle \bs y,\bs y' \rangle$ when $\bs \lambda=\bs 1_K:=(1,1,\ldots,1)\in\mathbb{R}^K$. WIP also reduces to the indefinite inner product  $\langle \bs y_i^+,\bs y_j^+ \rangle - \langle \bs y_i^-,\bs y_j^- \rangle$ when $\bs \lambda=(1,\ldots,1,-1,\ldots,-1) = (\bs 1_{K-q}, -\bs 1_{q})\in\mathbb{R}^K$.

For alleviating the problem to choose an appropriate value of $q$ in IPDS, we specify WIP $g_{\bs \lambda}(\bs y, \bs y') = \langle \bs y,\bs y' \rangle_{\bs \lambda}$ in the model (\ref{eq:parametrized_similarity}).
We then propose \textit{weighted inner product similarity} (WIPS) defined by
\begin{align*}
h_{\text{WIPS}}(\bs x, \bs x';\bs \theta,\bs \lambda) 
&= \langle \bs f_{\bs \theta}(\bs x),\bs f_{\bs \theta}(\bs x') \rangle_{\bs \lambda},
\end{align*}
where $\bs f_{\bs \theta}:\mathbb{R}^p \to \mathbb{R}^K$ is a non-linear transformation parameterized by $\bs \theta\in \bs \Theta$, and $\bs \lambda\in\mathbb{R}^K$ is also a parameter vector to be estimated.
The parameter vectors $\bs \theta$ and $\bs \lambda$ are jointly optimized in the objective function (\ref{eq:logistic_loss}).
WIPS obviously includes IPS, SIPS, and IPDS as special cases by specifying $\bs\lambda$ and $\bs f_{\bs \theta}$  (Table~\ref{table:correspondence}).
WIPS inherits the universal approximation capability of general similarities from IPDS, but
WIPS does \emph{not} extend IPDS for a larger class of similarity functions;
WIPS is expressed as IPDS by redefining $\bs f_{\bs \theta}$ so that $k$-th element is rescaled by $\sqrt{\lambda_k}$ for $\lambda_k>0$  and $\sqrt{-\lambda_k}$ for $\lambda_k<0$.
However, the discrete optimization of $0\le q\le K$ in IPDS is mitigated by the continuous optimization of $\bs \lambda\in\mathbb{R}^K$ in WIPS.

\begin{table}[htbp]
\centering
\begin{tabular}{lcc}
\toprule[0.2ex]
WIPS & $\bs \lambda$ & $\bs f_{\bs \theta}$ \\
\cmidrule(lr){1-3}
IPS & $\bs 1_K$ & $\bs f_{\bs \theta}$ \\
SIPS & $(\bs 1_{K+1},-1)$ & $(\tilde{\bs f}_{\bs \theta}(\bs x),u_{\bs \theta}(\bs x),1,u_{\bs \theta}(\bs x)-1)$ \\
IPDS & $(\bs 1_{K-q},-\bs 1_q)$ & $(\bs f_{\bs \theta}^+(\bs x),\bs f_{\bs \theta}^-(\bs x))$\\
\bottomrule[0.2ex]
\end{tabular}
\caption{WIPS expresses the other models by specifying $\bs \lambda$ and $\bs f_{\bs \theta}$. }
\label{table:correspondence}
\end{table}

\subsection{Interpreting WIPS as a Matrix Decomposition}

Let $h_*$ be a general similarity function that is not limited to PD or CPD similarities.
WIPS can be interpreted as the eigen-decomposition of the similarity matrix $\bs H_*$ whose $(i,j)$-th entry is $h_*(\bs x_i,\bs x_j)$.
Since $\bs H_*$ is a symmetric real-valued matrix, it can be expressed as
\[
    \bs H_*= \bs U \bs \Lambda \bs U^\top,
\]
where $\bs U = (\bs u_1,\ldots, \bs u_n) \in \mathbb{R}^{n\times n}$ is the orthogonal matrix of unit eigen-vectors and and $\bs \Lambda = \text{diag}(\lambda_1,\ldots,\lambda_n)$ is the diagonal matrix of real eigenvalues $\lambda_k\in\mathbb{R}$.
There will be negative $\lambda_k$'s when $h^*$ is a non-PD similarity.
Let us reorder the eigenvalues so that $|\lambda_1| \ge |\lambda_2| \ge \cdots \ge |\lambda_n|$, and take the first $K$ eigen-vectors as $\bs U_K = (\bs u_1,\ldots, \bs u_K) \in \mathbb{R}^{n\times K}$ and also $K$ eigenvalues as $\bs \Lambda_K = \text{diag}(\bs \lambda)$ with $\bs \lambda = (\lambda_1,\ldots,\lambda_K)$.
Now the similarity matrix is approximated as
\[
 \bs H_* \approx \bs U_K \bs \Lambda_K \bs U_K^\top.
\]
We may define feature vectors $\bs y_i\in\mathbb{R}^K$ for $\bs x_i$, $i=1,\ldots,n$ as $\bs U_K^\top = (\bs y_1,\ldots, \bs y_n)$. This results in
\begin{equation}
 h_*(\bs x_i,\bs x_j) \approx \langle \bs y_i, \bs y_j \rangle_{\bs \lambda}
 \label{eq:happroxwips}
\end{equation}
for all $1\le i < j \le n$,
and the accuracy improves in (\ref{eq:happroxwips}) as $K$ approaches $n$.

The representation learning with WIPS can be interpreted as a relaxed version of the matrix decomposition without any constraints such as the orthogonality of vectors. Instead, we substitute $\bs y_i = \bs f_{\bs \theta}(\bs x_i)$ in (\ref{eq:happroxwips}) and optimize the parameter vectors $\bs \theta$ and $\bs \lambda$ with an efficient algorithm such as SGD.
The above interpretation gives an insight into how WIPS deals with general similarities; allowing negative $\lambda_k$ in WIPS is essential for non-PD similarities.

\section{Experiments} 

\subsection{Preliminary}
\paragraph{Datasets.}
We use the following networks and a text corpus.

\begin{itemize}
\item WebKB Hypertext Network\footnote{\url{https://linqs.soe.ucsc.edu/data}} has 877 nodes and 1,480 links. Each node represents a hypertext. A link between nodes represents their hyperlink ignoring the direction.
Each node has a data vector of 1,703 dimensional bag-of-words.
Each node also has semantic class label which is one of \{Student, Faculty, Staff, Course, Project\}, and university label which is one of \{Cornell, Texas, Washington, Wisconsin\}.

\item DBLP Co-authorship Network~\cite{prado2013mining} has 41,328 nodes and 210,320 links. Each node represents an author. A link between nodes represents any collaborations.
Each node has a 33 dimensional data vector that represents the number of publications in each of 29 conferences and journals, and 4 microscopic topological properties describing the neighborhood of the node.

\item WordNet Taxonomy Tree~\cite{NIPS2017_7213poincare} has 37,623 nodes and 312,885 links. Each node represents a word. A link between nodes represents a hyponymy-hypernymy relation.  
Each node has a 300 dimensional data vector prepared from Google's pre-trained word embeddings\footnote{\url{https://code.google.com/archive/p/word2vec}}.

\item Text9 Wikipedia Corpus\footnote{\label{text9}Extension of Text8 corpus, \url{http://mattmahoney.net/dc/textdata}} consists of English Wikipedia articles containing about 123M tokens.
\end{itemize}

\paragraph{Similarity models.}
We compare WIPS with baseline similarity models IPS, SIPS and IPDS~\cite{DBLP:conf/icml/OkunoHS18,DBLP:journals/corr/abs-1810-03463} in addition to the negative
Poincar\'e distance~\cite{NIPS2017_7213poincare}.
Because most feature vectors obtained by Poincar\'e distance are close to the rim of Poincar\'e ball,
we also consider a natural representation in hyperbolic space by transformation $y_i\rightarrow 2y_i/(1-||\bs y||_2^2)$ for downstream tasks (denoted as Hyperbolic in Results).
We focus only on comparing the similarity models, because some experiments~\cite{DBLP:conf/icml/OkunoHS18} showed that even the simplest IPS model with neural networks can outperform many existing feature learning methods, such as DeepWalk~\cite{perozzi2014deepwalk} or GraphSAGE~\cite{Hamilton2017InductiveRL}.

\paragraph{NN architecture.}
For computing feature vectors, we use a 2-hidden layer fully-connected neural network $\bs f_{\bs \theta}:\mathbb{R}^p \to \mathbb{R}^K$ where each hidden layer consists of 2,000 hidden units with ReLU activation function.

\paragraph{Model training.}
The neural network models are trained by an Adam optimizer with negative sampling approach~\cite{Mikolov:2013:DRW:2999792.2999959,Tang:2015:LLI:2736277.2741093} whose batch size is 64 and the number of negative sample is 5.
The initial learning rate is grid searched over \{2e-4, 1e-3\}.
The dimensionality ratio $q/K$ of IPDS is grid searched over $\{0.01,0.25,0.5,0.75,0.99\}$.
Instead, $\bs \lambda$ in WIPS is initialized randomly from the uniform distribution of range (0, $1/K$).

\paragraph{Implementations.}
To make fair ``apples-to-apples'' comparisons, 
all similarity models are implemented with PyTorch framework upon the SIPS library\footnote{\url{https://github.com/kdrl/SIPS}}.
We also made C implementation upon the word2vec program\footnote{\url{https://github.com/tmikolov/word2vec}}.
All our implementations are publicly available at GitHub\footnote{\url{https://github.com/kdrl/WIPS}}.

\begin{figure*}[!t]
\centering
\noindent\includegraphics[width=\textwidth]{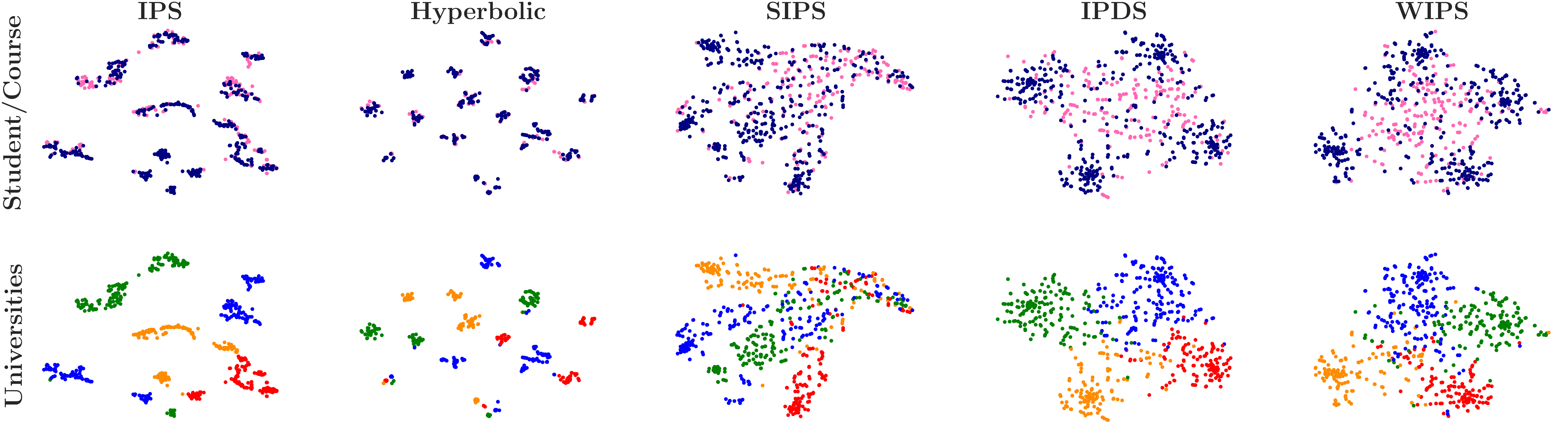} 
\caption{Visualization of Hypertext Network.
Two major semantic classes, Student (415 nodes) and Course (218 nodes), are plotted.
10 dimensional feature vectors are mapped to 2d space with T-SNE.
The nodes are colored by its semantic labels (upper) for Student (navy) and Course (pink), and also university labels (lower) for Cornell (red), Texas (orange), Washington (green) and Wisconsin (blue).
Both class labels are clearly identified with IPDS and WIPS, whereas they become obscure in the other embeddings.
} \label{fig:visualization}
\end{figure*}

\subsection{Evaluation Tasks}
To assess both the approximation ability of the proposed similarity model as well as the effectiveness of the learned feature vectors,
we conduct the following tasks.

\paragraph{Reconstruction (Task 1).}
Most settings are the same as those in Okuno et al.~\shortcite{DBLP:journals/corr/abs-1810-03463}.
To evaluate approximation capacity, we embed all nodes and predict all links, i.e., reconstruct the graph structure from the learned feature vectors.
The maximum number of iterations for training is set to 100K.
ROC-AUC of reconstruction error is evaluated at every 5K iterations and the best score is reported for each model.

\paragraph{Link prediction (Task 2).}
To evaluate generalization error, we randomly split the nodes into train (64\%), validation (16\%) and test (20\%) sets with their associated links.
Models are trained on the sub-graph of train set and evaluated occasionally on validation set in iterations; the best evaluated model is selected.
The maximum number of iterations and validation interval are 5K/100 for Hypertext Network, and they are 40K/2K for Co-authorship Network and Taxonomy Tree.
Then, the trained models are evaluated by predicting links from unseen nodes of test set and ROC-AUC of the prediction error is computed.
This is repeated for 10 times and the average score is reported for each model.

\begin{table}[!t]
\centering
\begin{adjustbox}{width=\linewidth}
\begin{tabular}{clcccccc}

\toprule[0.2ex]
 & & \multicolumn{6}{c}{\textbf{Dimensionality}}\\
\cmidrule(lr){3-8}
 & & \multicolumn{3}{c}{\textbf{Reconstruction}} & \multicolumn{3}{c}{\textbf{Link prediction}} \\
\cmidrule(lr){3-5}\cmidrule(lr){6-8}
 & & 10 & 50 & 100 & 10 & 50 & 100 \\
\midrule[0.1ex]

\multirow{5}{*}{\shortstack[l]{
 \rotatebox[origin=c]{90}{\textbf{Hypertext}}
}}
  & IPS             & 91.99        &       94.23  &       94.24  &       77.73  &       77.62  &       77.16 \\ 
  & Poincar\'e      & 94.09        &       94.13  &       94.11  & \sigu{82.21} &       79.64  &       79.48 \\ 
  & {SIPS}          & \sigu{95.11} & \sigu{95.12} & \sigu{95.12} &      {82.01} & \sigu{81.84} & \sigu{81.13}\\ 
  & {IPDS}          & \sigb{95.12} & \sigu{95.12} & \sigb{95.12} & \sigb{82.59} & \sigb{82.75} & \sigu{82.19}\\ 
  & \textbf{WIPS}   & \sigu{95.11} & \sigb{95.12} & \sigu{95.12} & \sigu{82.38} & \sigu{82.68} & \sigb{82.93}\\
\cmidrule(lr){1-8}
\multirow{5}{*}{\shortstack[l]{
 \rotatebox[origin=c]{90}{\textbf{Co-author}}
}}
   & IPS            &       85.01  &       86.02  &       85.80  &       83.83  &       84.41  &       84.02 \\
   & Poincar\'e     &       86.84  &       86.69  &       86.72  &       85.82  &       85.92  &       85.93 \\  
   & {SIPS}         & \sigu{90.01} & \sigu{91.35} & \sigu{91.06} & \sigu{88.24} & \sigu{88.69} & \sigu{88.67}\\  
   & {IPDS}         & \sigu{90.13} & \sigu{91.68} & \sigu{91.59} & \sigb{88.42} & \sigu{88.97} & \sigu{88.85}\\  
   & \textbf{WIPS}  & \sigb{90.50} & \sigb{92.44} & \sigb{92.95} & \sigu{88.16} & \sigb{89.43} & \sigb{89.40}\\  
\cmidrule(lr){1-8}
\multirow{5}{*}{\shortstack[l]{
 \rotatebox[origin=c]{90}{\textbf{Taxonomy}}
}}
   & IPS            & 79.95        &       75.80  &       74.97  &       67.25  &       65.71  &      65.38 \\
   & Poincar\'e     & 91.69        &       89.10  &       88.97  &       83.04  &       79.52  &      78.97 \\
   & {SIPS}         & \sigu{98.78} & \sigu{99.75} & \sigu{99.77} & \sigu{90.42} & \sigu{92.12} &\sigu{92.09}\\
   & {IPDS}         & \sigb{99.65} & \sigb{99.89} & \sigb{99.90} & \sigb{95.99} & \sigb{96.37} &\sigu{96.41}\\
   & \textbf{WIPS}  & \sigu{99.64} & \sigu{99.85} & \sigu{99.87} & \sigu{95.07} & \sigu{96.36} &\sigb{96.51}\\
\bottomrule[0.2ex]

\end{tabular}
\end{adjustbox}
\caption{ROC-AUC for the reconstruction from fully observed data (\textbf{Task 1}) and the inductive link prediction of unseen nodes (\textbf{Task 2}). Boldface is the best, and underlines are 2nd and 3rd scores.}
\label{table:task1andtask2}
\end{table}

\begin{table}[!t]
\centering
\begin{adjustbox}{width=\linewidth}
 \begin{tabular}{lcccccc}
  \toprule[0.2ex]
   & IPS & Poincar\'e & Hyperbolic &SIPS &  IPDS & WIPS\\
  \midrule[0.1ex]
  A &56.08&46.19&47.22&\sigu{69.09}&\sigu{71.70}&\sigb{73.35}\\
  B &91.59&30.17&93.12&\sigu{93.81}&\sigu{93.81}&\sigb{96.31}\\
  \bottomrule[0.2ex]
 \end{tabular}
\end{adjustbox}
\caption{Classification accuracies on predicting (A) the semantic class label and (B) the university label (\textbf{Task 3}).} \label{table:classification}
\end{table}

\paragraph{Hypertext classification (Task 3).}
To see the effectiveness of learned feature vectors,
we conduct hypertext classification with inductive setting.
First, feature vectors of all nodes are computed by the models trained for Hypertext Network in Task~2,
where $K$ is also optimized from $\{10, 50, 100\}$.
Then, a logistic regression classifier is trained on the feature vectors of train set for predicting class labels.
The classification accuracy is measured on test set.
This is repeated for 10 times and the average score is reported for each model.

\paragraph{Word similarity (Task 4).}
The similarity models are applied to Skip-gram model~\cite{Mikolov:2013:DRW:2999792.2999959} to learn word embeddings.
To see the quality of the learned word embeddings from Text9 corpus,
word-to-word similarity scores induced by the word embeddings are compared with four human annotated similarity scores: SimLex-999~\cite{Hill:2015:SES:2893320.2893324}, YP-130~\cite{Yang:2005:MSS:1082161.1082196}, WS-Sim and WS-Rel~\cite{Agirre:2009:SSR:1620754.1620758}.
The quality is evaluated by Spearman's rank correlation.
The models are trained with the SGD optimizer used in Mikolov et al.~\shortcite{Mikolov:2013:DRW:2999792.2999959}.
We set the number of negative samples to 5 and the number of epochs to 5.
The initial learning rate is searched over $\{0.01, 0.025\}$.
As additional baselines, we use the original Skip-gram (SG)~\cite{Mikolov:2013:DRW:2999792.2999959} and Hyperbolic Skip-gram (HSG)~\cite{leimeister2018skipgram}.
Since SG uses twice as many parameters, we also compare SG with $K/2$ dimensional word vectors for fair comparisons.
To compute similarities with SG, both inner product and cosine similarity (denoted as SG and SG$^*$) are examined.
Each model is trained 3 times and the average score is reported.

\begin{table}[!t]
\centering
\begin{adjustbox}{width=\linewidth}
 \begin{tabular}{lccccccccccc}
  \toprule[0.2ex]
& \multicolumn{2}{c}{\textbf{SimLex}} & \multicolumn{2}{c}{\textbf{YP}} & \multicolumn{2}{c}{\textbf{WS$_{\text{SIM}}$}} & \multicolumn{2}{c}{\textbf{WS$_{\text{REL}}$}}\\
\cmidrule(lr){2-3}\cmidrule(lr){4-5}\cmidrule(lr){6-7}\cmidrule(lr){8-9}
    & 10 & 100 & 10 & 100 & 10 & 100  & 10 & 100 \\
  \midrule[0.1ex]
  IPS               &13.6&23.6&17.5&37.3&46.0&73.8&42.3&69.8\\
  SIPS              &17.1&31.1&24.9&\sigu{48.0}&55.9&\sigu{77.0}&\sigu{49.8}&\sigu{71.2}\\
  IPDS              &16.9&\sigu{31.3}&\sigu{25.7}&\sigu{48.9}&\sigu{56.2}&\sigu{76.8}&\sigb{49.9}&\sigu{71.4}\\
  WIPS              &\sigu{19.2}&\sigb{31.4}&\sigu{27.2}&\sigb{49.0}&\sigb{57.0}&\sigb{78.0}&\sigu{48.7}&\sigb{71.5}\\
\cmidrule(lr){1-9}
  SG(K/2)           &15.6&27.5&9.90&23.8&20.7&69.1&28.9&67.0\\
  SG$^{*}$(K/2)     &17.0&27.8&18.2&36.4&43.3&75.7&27.1&65.2\\
  SG                &\sigu{18.6}&30.9&14.1&31.0&46.1&71.5&46.4&68.7\\
  SG$^{*}$          &\sigb{20.9}&\sigu{31.3}&\sigb{27.3}&39.3&\sigu{56.3}&75.4&39.7&67.1\\
  HSG               &19.3&25.8&23.5&39.6&52.9&68.2&36.1&58.2\\
  \bottomrule[0.2ex]
 \end{tabular}
\end{adjustbox}
\caption{Spearman's $\rho$ on the word similarity task (\textbf{Task 4}).}\label{table:spearman}
\end{table}

\subsection{Results}
Looking at the reconstruction error in Table~\ref{table:task1andtask2}, the approximation capabilities of IPDS and WIPS are very high, followed by SIPS, Poincar\'e and IPS as expected from the theory.
This tendency is also seen in the visualization of Fig.~\ref{fig:visualization}.
Looking at the link prediction error in Table~\ref{table:task1andtask2}, the difference of models is less clear for generalization performance, but IPDS and WIPS still outperform the others.
Table~\ref{table:classification} shows the effectiveness of the feature vectors, where IPDS and WIPS have again the best or almost the best scores.
As expected, the hyperbolic representation is much better than Poincar\'e ball model (Note: this does not apply to Table~\ref{table:task1andtask2}, because these two representations are in fact the same similarity model).
Table~\ref{table:spearman} shows the quality of word embeddings.
The new similarity models SIPS, IPDS and WIPS work well with similar performance to SG$^*$ (with twice the number of parameters), whereas IPS works poorly for most cases.

As expected, we can see that IPDS and WIPS are very similar in all the results.
Although IPDS and WIPS have the same level of approximation capability, WIPS is much easier to train.
For example, the total training time for WIPS was much shorter than IPDS in the experiments.
In Task 1 with Hypertext network and $K=100$, IPDS took about 110 min to achieve 95.12 AUC with the five searching points for tuning the dimensionality ratio. But, WIPS only took about 20 min and achieved 95.12.
While WIPS stably optimizes the dimensionality ratio of the two IPS models, IPDS is sensitive to the value of dimensionality ratio.
For instance, in Task 2 with Hypertext network and $K=100$, we observed that AUC performance of IPDS varies from the worst 64.31 to the best 82.58; WIPS was 82.93.

One of the main reason to use expressive similarity model is to skip the model selection phase, which is often very time-consuming.
However, if the training of the model is difficult, the benefit has no meaning and the practitioner needs to take much time and effort.
Through the experiments, we can see that WIPS is useful in that it is not only as expressive as IPDS but also easy to train which is very important in practice.

\section{Conclusion}
We proposed a new similarity model WIPS by introducing tunable element-wise weights in inner product.
Allowing positive and negative weight values, it has been shown theoretically and experimentally that WIPS has the same approximation capability as IPDS, which has been proved to universally approximate general similarities.
As future work, we may use WIPS as a building block of larger neural networks, or extend WIPS to tensor such as $\sum_{k=1}^K w_k y_k y_k' y_k''$ to represent relations of three (or more) feature vectors $\bs y, \bs y', \bs y''$.
This kind of consideration is obviously not new, but our work gives theoretical and practical insight to neural network architecture.

\section*{Acknowledgements}
We would like to thank anonymous reviewers for their helpful advice.
This work was partially supported by JSPS KAKENHI grant 16H02789 to HS, 17J03623 to AO and 18J15053 to KF.

\bibliographystyle{named}

\end{document}